\documentclass[11pt,a4paper]{article}
\usepackage[hyperref]{acl2019}
\usepackage{times}
\usepackage{latexsym}
\usepackage{amsmath}

\usepackage{epsfig}
\usepackage{graphicx}
\usepackage{amsmath}
\usepackage{amssymb}
\usepackage{float}
\usepackage{multirow}
\usepackage{color}
\usepackage{graphics}
\usepackage{comment}
\usepackage{url}
\usepackage{textcomp}
\usepackage{mathtools}

\usepackage{url}

\aclfinalcopy 
\def\aclpaperid{2119} 

\title{Attention Is (not) All You Need for Commonsense Reasoning}

\author{{Tassilo Klein$^1$, Moin Nabi$^1$} \\
$^1$SAP Machine Learning Research, Berlin, Germany\\
{\tt $\{$tassilo.klein, m.nabi$\}$@sap.com}}

\date{}
\begin{document}
\maketitle
\begin{abstract}
The recently introduced BERT model exhibits strong performance on several language understanding benchmarks. 
In this paper, we describe a simple re-implementation of BERT for commonsense reasoning. We show that the attentions produced by BERT can be directly utilized for tasks such as the \emph{Pronoun Disambiguation Problem} and \emph{Winograd Schema Challenge}. Our proposed attention-guided commonsense reasoning method is conceptually simple yet empirically powerful. Experimental analysis on multiple datasets demonstrates that our proposed system performs remarkably well on all cases while outperforming the previously reported state of the art by a margin.  While results suggest that BERT seems to implicitly learn to establish complex relationships between entities, solving commonsense reasoning tasks might require more than unsupervised models learned from huge text corpora.
\end{abstract}

\section{Introduction}

Recently, neural models pre-trained on a language modeling task, such as ELMo \cite{peters2018deep}, OpenAI GPT \cite{radford2018improving}, and BERT~\cite{devlin2018bert}, have achieved impressive results on various natural language processing tasks such as question-answering and natural language inference. 
 The success of BERT can largely be associated to the notion of context-aware word embeddings, which differentiate it from common approaches such as word2vec~\cite{DBLP:journals/corr/abs-1301-3781} that establish a static semantic embedding. 
Since the introduction of BERT, the NLP community continues to be impressed by the amount of ideas produced on top of this powerful language representation model. However, despite its success, it remains unclear whether the representations produced by BERT can be utilized for tasks such as commonsense reasoning. Particularly, it is not clear whether BERT shed light on solving tasks such as the \emph{Pronoun Disambiguation Problem}~(PDP) and \emph{Winograd Schema Challenge}~(WSC). These tasks have been proposed as potential alternatives to the Turing Test, because they are formulated to be robust to statistics of word co-occurrence~\cite{levesque2012winograd}. 

Below is a popular example from the binary-choice pronoun coreference problem \cite{lee2017end} of WSC: \\
\\\emph{\textbf{Sentence:}} \emph{The trophy doesn\textquotesingle t fit in the suitcase because {\textbf{it}} is too small.}\\ 
\emph{\textbf{Answers:}} \emph{\textbf{A)}} the trophy \emph{\textbf{B)}} the suitcase\\ 

Humans resolve the pronoun ``it'' to ``the suitcase'' with no difficulty, whereas a system without commonsense reasoning would be unable to distinguish ``the suitcase'' from the otherwise viable candidate, ``the trophy''. 

\begin{figure*}[t!]
\begin{center}
\includegraphics[width=0.85\textwidth]{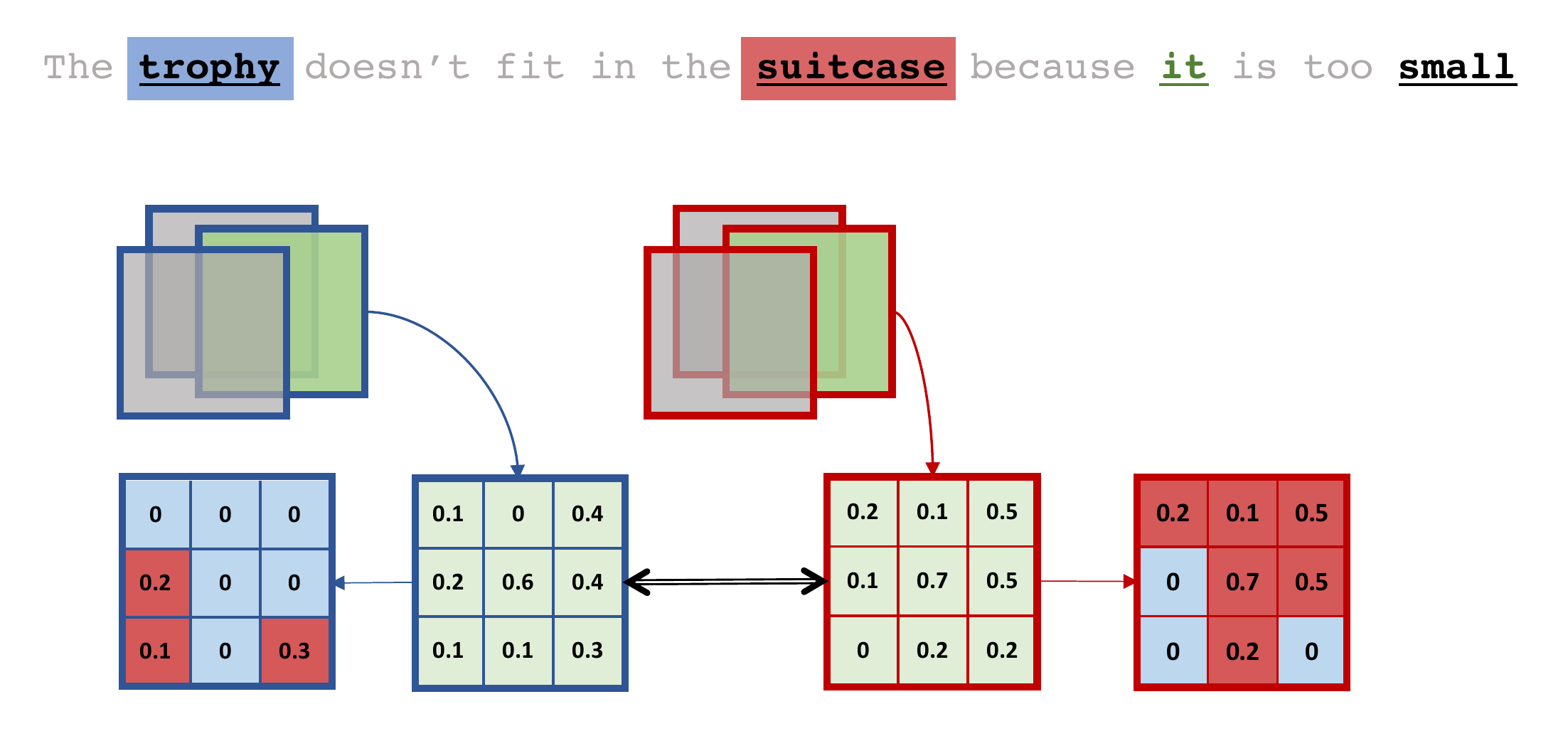}
\end{center}
\caption{Maximum Attention Score (MAS) for a particular sentence, where colors show attention maps for different words (best shown in color). Squares with blue/red frames correspond to specific sliced attentions $A_c$ for candidates $c$, establishing the relationship to the reference pronoun indicated with green. Attention is color-coded in blue/ red for candidates ``trophy''/ ``suitcase''; the associated pronoun ``it'' is indicated in green. Attention values are compared elementwise (black double arrow), and retain only the maximum achieved by a masking operation. Matrices on the outside with red background elements correspond to the masked attentions $A_c \circ M_c$.}\label{fig:cat}
\end{figure*}

Previous attempts at solving WSC usually involve heavy utilization of annotated knowledge bases (KB), rule-based reasoning, or hand-crafted features \cite{peng2015solving, bailey2015winograd, schuller2014tackling, sharma2015towards, morgenstern2016planning}. There are also some empirical works towards solving WSC making use of learning \cite{rahman2012resolving, tang2018self, radford2018improving}. Recently, \cite{trinh2018simple} proposed to use a language model (LM) to score the two sentences obtained when replacing the pronoun by the two candidates. The sentence that is assigned higher probability under the model designates the chosen candidate. Probability is calculated via the chain rule, as the product of the probabilities assigned to each word in the sentence. Very recently, \cite{emami2018knowledge} proposed the knowledge hunting method, which is a rule-based system that uses search engines to gather evidence for the candidate resolutions without relying on the entities themselves. Although these methods are interesting, they need fine-tuning, or explicit substitution or heuristic-based rules. See also ~\cite{trichelair2018evaluation} for a discussion.

The BERT model is based on the ``Transformer'' architecture~\cite{vaswani2017attention}, which relies purely on attention mechanisms, and does not have an explicit notion of word order beyond marking each word with its absolute-position embedding. This reliance on attention may lead one to expect decreased performance on commonsense reasoning tasks~\cite{roemmele2011choice, zellers2018swag} compared to RNN (LSTM) models~\cite{hochreiter1997long} that do model word order directly, and explicitly track states across the sentence. However, the work of~\cite{peters-etal-2018-dissecting} suggests that bidirectional language models such as BERT implicitly capture some notion of coreference resolution.

\begin{table*}[t!]
\begin{center}
\centering
\begin{tabular}{ll}
\hline
Method & Accuracy \\
\hline\hline
Unsupervised Semantic Similarity Method (USSM) & 48.3 \% \\
USSM + Cause-Effect Knowledge Base \cite{liu2016probabilistic} & 55.0 \% \\
USSM + Cause-Effect + WordNet \cite{miller1995wordnet} + ConceptNet \cite{liu2004conceptnet} KB & 56.7 \% \\
Subword-level Transformer LM ~\cite{vaswani2017attention} & 58.3 \% \\
Single LM (partial) \cite{trinh2018simple}& 53.3 \% \\
Single LM (full) \cite{trinh2018simple}& 60.0 \% \\
\hline\hline
Patric Dhondt (WS Challenge 2016) & 45.0 \% \\
Nicos Issak (WS Challenge 2016) & 48.3 \% \\
Quan Liu (WS Challenge 2016 - \textbf{winner}) & 58.3 \% \\
USSM + Supervised Deepnet & 53.3 \% \\
USSM + Supervised Deepnet + 3 Knowledge Bases & 66.7 \% \\
\hline\hline
\textbf{Our Proposed Method} & \textbf{68.3 \%}\\
\hline
\end{tabular}
\caption{Pronoun Disambiguation Problem: Results on (top) Unsupervised method performance on PDP-60 and (bottom) Supervised method performance on PDP-60. Results other than ours are taken from \cite{trinh2018simple}.}
\label{tab:results-pdp}
\end{center}
\end{table*}
\begin{table}
\begin{center}
\centering
\begin{tabular}{ll}
\hline
Method & Accuracy \\
\hline\hline
Random guess & 50.0 \% \\
USSM + KB & 52.0\% \\
USSM + Supervised DeepNet + KB & 52.8 \% \\
Single LM \cite{trinh2018simple}& 54.5 \% \\
Transformer ~\cite{vaswani2017attention} & 54.1 \% \\
Know. Hunter \cite{emami2018knowledge}& 57.1 \% \\
\hline\hline
\textbf{Our Proposed Method} & \textbf{60.3 \%}\\
\hline
\end{tabular}
\caption{Results for Winograd Schema Challenge. The other results are taken from \cite{trichelair2018evaluation} and \cite{trinh2018simple}.}
\label{tab:results-win}
\end{center}
\end{table} 
In this paper, we show that the attention maps created by an out-of-the-box BERT can be directly exploited to resolve coreferences in long sentences. As such, they can be simply repurposed for the sake of commonsense reasoning tasks while achieving state-of-the-art results on the multiple task. On both PDP and WSC, our method outperforms previous state-of-the-art methods, without using expensive annotated knowledge bases or hand-engineered features. On a Pronoun Disambiguation dataset, PDP-60, our method achieves 68.3\% accuracy, which is better than the state-of-art accuracy of 66.7\%. On a WSC dataset, WSC-273, our method achieves 60.3\%. As of today, state-of-the-art accuracy on the WSC-273 for single model performance is around 57\%, \cite{emami2018knowledge} and \cite{trinh2018simple}. These results suggest that BERT implicitly learns to establish complex relationships between entities such as coreference resolution. Although this helps in commonsense reasoning, solving this task requires more than employing a language model learned from large text corpora.

\section{Attention Guided Reasoning}
In this section we first review the main aspects of the BERT approach, which are important to understand our proposal and we introduce notations used in the rest of the paper. Then, we introduce Maximum Attention Score (MAS), and explain how it can be utilized for commonsense reasoning.

\subsection{BERT and Notation}
The concept of BERT is built upon two key ingredients: (a) the transformer architecture and (b) unsupervised pre-training. 

The transformer architecture consists of two main building blocks, stacked encoders and decoders, which are connected in a cascaded fashion. The encoder is further divided into two components, namely a self-attention layer and a feed-forward neural network. The self-attention allows for attending to specific words during encoding and therefore establishing a focus context w.r.t. to each word. In contrast to that, the decoder has an additional encoder-decoder layer that switches between self-attention and a feed-forward network. It allows the decoder to attend to specific parts of the input sequence. As attention allows for establishing a relationship between words, it is very important for tasks such as coreference resolution and finding associations. In the specific context of pronouns, attention gives rise to links to $m$ candidate nouns, which we denote in the following as $\mathcal{C}=\left\{c_1,..,c_m\right\}$.
The concept of self-attention is further expanded within BERT by the idea of so called multi-head outputs that are incorporated in each layer. In the following, we will denote heads and layers with $h\in H$ and $l\in L$, respectively. Multi-heads serve several purposes. On the one hand, they allow for dispersing the focus on multiple positions. On the other hand, they constitute an enriched representation by expanding the embedding space.
Leveraging the nearly unlimited amount of data available, BERT learns two novel unsupervised prediction tasks during training. One of the tasks is to predict tokens that were randomly masked given the context, notably with the context being established in a bi-directional manner. The second task constitutes next sentence prediction, whereby BERT learns the relationship between two sentences, and classifies whether they are consecutive.  

\begin{figure*}[t!]
\begin{center}
\includegraphics[width=0.95\textwidth]{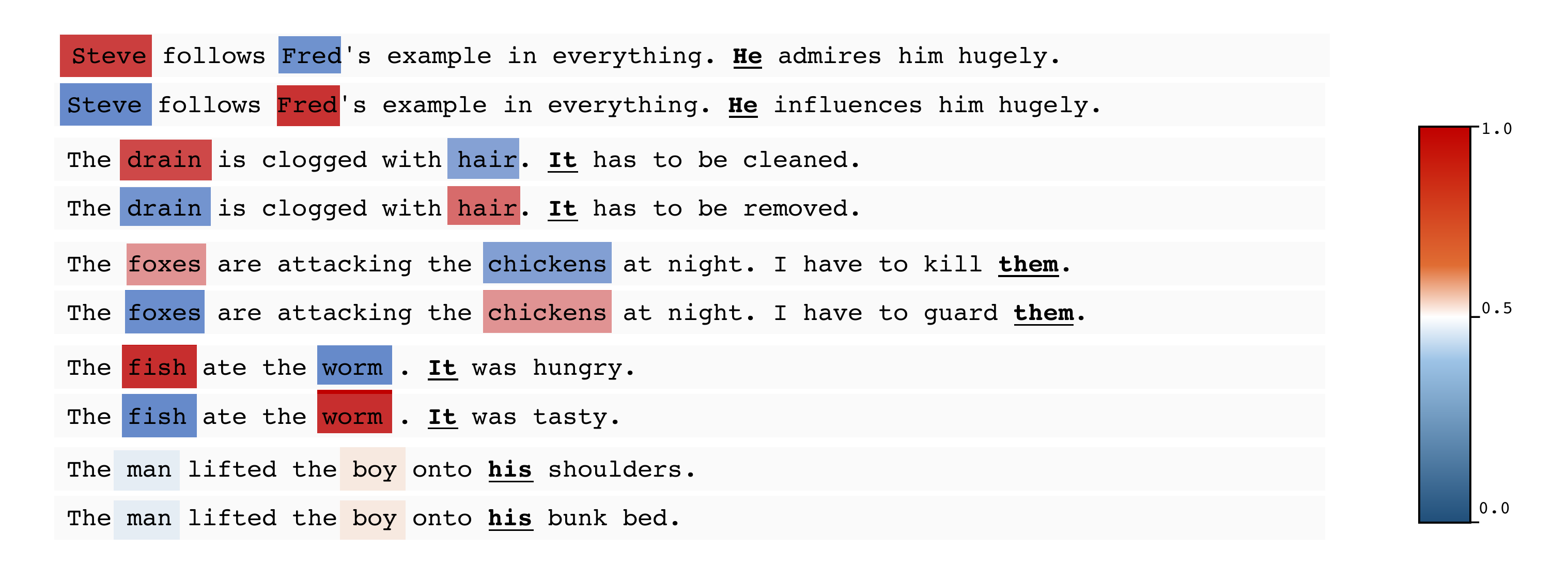}
\end{center}
\caption{Maximum Attention Score (MAS) for some sample questions from WSC-273: The last example is an example of failure of the method, where the coreference is predicted incorrectly.}\label{fig:fig_2}
\end{figure*}

\subsection{Maximum Attention Score (MAS)}
In order to exploit the associative leverage of self-attention, the computation of MAS follows the notion of max-pooling on attention level between a reference word $s$ (e.g. pronoun) and candidate words $c$ (e.g. multiple choice pronouns). The proposed approach takes as input the BERT attention tensor and produces for each candidate word a score, which indicates the strength of association. To this end, the BERT attention tensor $A\in\mathbb{R}^{H\times L \times\mid\mathcal{C}\mid}$ is sliced into several matrices $A_c\in\mathbb{R}^{H\times L}$, each of them corresponding to the attention between the reference word and a candidate $c$. Each $A_c$ is associated with a binary mask matrix $M_c$. 
The mask values of $M_c$ are obtained at each location tuple $\left(l,h\right)$, according to:
\begin{equation}
M_{c}(l,h)=
\begin{dcases}
      1 & \operatorname*{argmax}A(l,h)=c \\
      0 & \text{otherwise} \\
\end{dcases}
\end{equation}
Mask entries are non-zero only at locations where the candidate word $c$ is associated with maximum attention. Limiting the impact of attention by masking allows to accommodate for the most salient parts.
Given the $A_c$ and $M_c$ matrix pair for each candidate $c$, the MAS can be computed. For this purpose, the sum of the Hadamard product for each pair is calculated first. Next, the actual score is obtained by computing the ratio of each Hadamard sum w.r.t. all others according to,
\begin{equation}
MAS(c)=\frac{\sum_{l,h}A_c \circ M_c }{\sum_{c \in \mathcal{C}} \sum_{l,h}A_c \circ M_c} \in \left[0,1\right].
\end{equation}
Thus MAS retains the attention of each candidate only where it is most dominant, coupling it with the notion of frequency of occurrence to weight the importance. See Fig.~\ref{fig:cat} for a schematic illustration of the computation of MAS, and the matrices involved.

\section{Experimental Results}
We evaluate our method on two commonsense reasoning tasks, PDP and WSC.

On the former task, we use the original set of 60 questions (PDP-60
) as the main benchmark. The second task (WSC-273
) is qualitatively much more difficult. The recent best reported result are not much above random guess. This task consists of 273 questions and is designed to work against traditional linguistic techniques, common heuristics or simple statistical tests over text corpora~\cite{levesque2012winograd}.

\subsection{BERT Model Details}
In all our experiments, we used the out-of-the-box BERT models without any task-specific fine-tuning. Specifically, we use the PyTorch implementation of pre-trained $bert-base-uncased$ models supplied by Google\footnote{https://github.com/huggingface/pytorch-pretrained-BERT}. This model has 12 layers (i.e., Transformer blocks), a hidden size of 768, and 12 self-attention heads. In all cases we set the feed-forward/filter size to be 3072 for the hidden size of 768. The total number of parameters of the model is 110M.
\subsection{Pronoun Disambiguation Problem}
We first examine our method on PDP-60 for the Pronoun Disambiguation task. In Tab. 1 (top), our method outperforms all previous unsupervised results sharply. Next, we allow other systems to take in necessary components to maximize their test performance. This includes making use of supervised training data that maps commonsense reasoning questions to their correct answer. As reported in Tab. 1 (bottom), our method outperforms the best system in the 2016 competition (58.3\%) by a large margin. Specifically, we achieve 68.3\% accuracy, better than the more recently reported results from \cite{liu2017combing} (66.7\%), who makes use of three KBs and a supervised deep network.
\subsection{Winograd Schema Challenge}
On the harder task WSC-273, our method also outperforms the current state-of-the-art, as shown in Tab. 2. Namely, our method achieves an accuracy of 60.3\%, nearly 3\% of accuracy above the previous best result. This is a drastic improvement considering the best system based on language models outperforms random guess by only 4\% in accuracy. 
This task is more difficult than PDP-60. First, the overall performance of all competing systems are much lower than that of PDP-60. Second, incorporating supervised learning and expensive annotated KBs to USSM provides insignificant gain this time (+3\%), comparing to the large gain on PDP-60 (+19\%). Finally, for the sake of completeness,~\cite{trinh2018simple} report that their  single language model trained on a customized dataset built from CommonCrawl based on questions used in comonsense reasoning achieves an higher accuracy than the proposed approach with 62.6\%.

We visualize the MAS to have more insights into the decisions of our resolvers. Fig.~\ref{fig:fig_2} displays some samples of correct and incorrect decisions made by our proposed method. MAS score of different words are indicated with colors, where the gradient from \emph{blue} to \emph{red} represents the score transition from low to high. 
\section{Discussion}

Pursuing commonsense reasoning in a purely unsupervised way seems very attractive for several reasons. On the one hand, this implies tapping the nearly unlimited resources of unannotated text and leveraging the wealth of information therein. On the other hand, tackling the commonsense reasoning objective in a (more) supervised fashion typically seems to boost performance for very a specific task as concurrent work shows~\cite{KCCYL2019}. However, the latter approach is unlikely to generalize well beyond this task. That is because covering the complete set of commonsense entities is at best extremely hard to achieve, if possible at all. The data-driven paradigm entails that the derived model can only make generalizations based on the data it has observed. Consequently, a supervised machine learning approach will have to be exposed to all combinations, i.e. replacing lexical items with semantically similar items in order to derive various concept notions. Generally, this is prohibitively expensive and therefore not viable. In contrast, in the proposed (unsupervised self-attention guided) approach this problem is alleviated. This can be largely attributed to the nearly unlimited text corpora on which the model originally learns, which makes it likely to cover a multitude of concept relations, and the fact that attention implicitly reduces the search space. However, all these approaches require the answer to explicitly exist in the text. That is, they are unable to resolve pronouns in light of abstract/implicit referrals that require background knowledge  - see~\cite{DBLP:journals/corr/abs-1810-00521} for more detail. However, this is beyond the task of WSC.
Last, the presented results suggest that BERT models the notion of complex relationship between entities, facilitating commonsense reasoning to a certain degree.

\section{Conclusion}
Attracted by the success of recently proposed language representation model BERT, in this paper, we introduce a simple yet effective re-implementation of BERT for commonsense reasoning. Specifically, we propose a method which exploits the attentions produced by BERT for the challenging tasks of \emph{PDP} and \emph{WSC}. The experimental analysis demonstrates that our proposed system outperforms the previous state of the art on multiple datasets. 
However, although BERT seems to implicitly establish complex relationships between entities facilitating tasks such as coreference resolution, the results also suggest that solving commonsense reasoning tasks might require more than leveraging a language model trained on huge text corpora. Future work will entail adaption of the attentions, to further improve the performance. 

\bibliography{acl2019}
\bibliographystyle{acl_natbib}

\end{document}